# Monte Carlo Pass Search for Setpieces

Simulating every delivery and set-up a team could have used, and how all 22 players respond

Andrew Kang

## Abstract

Set pieces produce about a third of non-penalty expected goals (xG), yet they are usually evaluated by what happened next. We present **Monte Carlo Pass Search (MCPS) for set pieces**, which treats a corner or a lofted free kick as a *distribution* over plausible executions and plausible futures instead of a single realised event. From the pre-kick state, MCPS proposes counterfactual deliveries and counterfactual set-ups (where the attackers stand at the kick), plays each out many times in a world model (3D ball physics, a generative model of all 22 players, the keeper's come-or-stay decision and a Player-to-Touch model) and scores every future with a value model. On 249 deliveries from 30 Serie A 2025/26 matches, with every model applied only to matches it never saw, the best available delivery is worth **0.0085 xG more** than the one kicked (+28 %), mainly because it reaches an attacker before the keeper. Searching where the attackers stand as well adds +0.0027 xG with the observed ball and +0.0033 on top of the best ball; designing their runs too adds +0.0007 more. We show how to keep such a search honest, and estimate how much more data each component needs before a club could rely on it.

## 1 Motivation

Set pieces decide matches. In our 30 matches, possessions that start from a corner or a free kick produce **33 % of non-penalty xG and 29 % of non-penalty goals**. They are also the most designable phase of play: the ball is dead, players walk into chosen positions, and one player chooses where to aim. Both choices, the delivery and the set-up, are decisions, and a decision should be judged against the alternatives that were available at the time.

Sports analytics is dominated by simpler methods that score the realised event: counts and rates (shots per corner, first-contact win rates), delivery-zone maps, and value models such as xG, expected threat (xT), VAEP and possession value (PV) [3, 4, 14]. They are cheap, transparent and useful, but they are **point estimates of what happened**, which limits them for set-piece decisions in three ways.

- **Outcome bias.** A team takes about 160 lofted set-piece deliveries a season. One header in or out flips a zone's record, and a good delivery that a defender happens to clear is scored as a bad one.
- **No counterfactuals.** Counts describe what was kicked, not what could have been. A zone nobody aims at, or a set-up nobody uses, has no data.
- **Interaction.** A delivery's value depends on how all 22 players respond: the attacker who times his run, the defender who follows him, the keeper who comes or stays. "What if the ball had gone 3 m further and lower?" cannot be read off past plays in which the players did something else.

MCPS takes a different view, adapted from open-play passing [7] (Figure 1). A delivery is a distribution over plausible kicks and plausible short-term futures. We borrow the search-and-evaluate idea of Monte Carlo Tree Search [12] with three parts: a **policy** that proposes counterfactual deliveries and set-ups, a **world model** that plays each one forward until the first touch, and a **value model** that scores the result. Averaging over futures gives each option's expected value; comparing options gives the best available choice and how much it beats the one that was made.

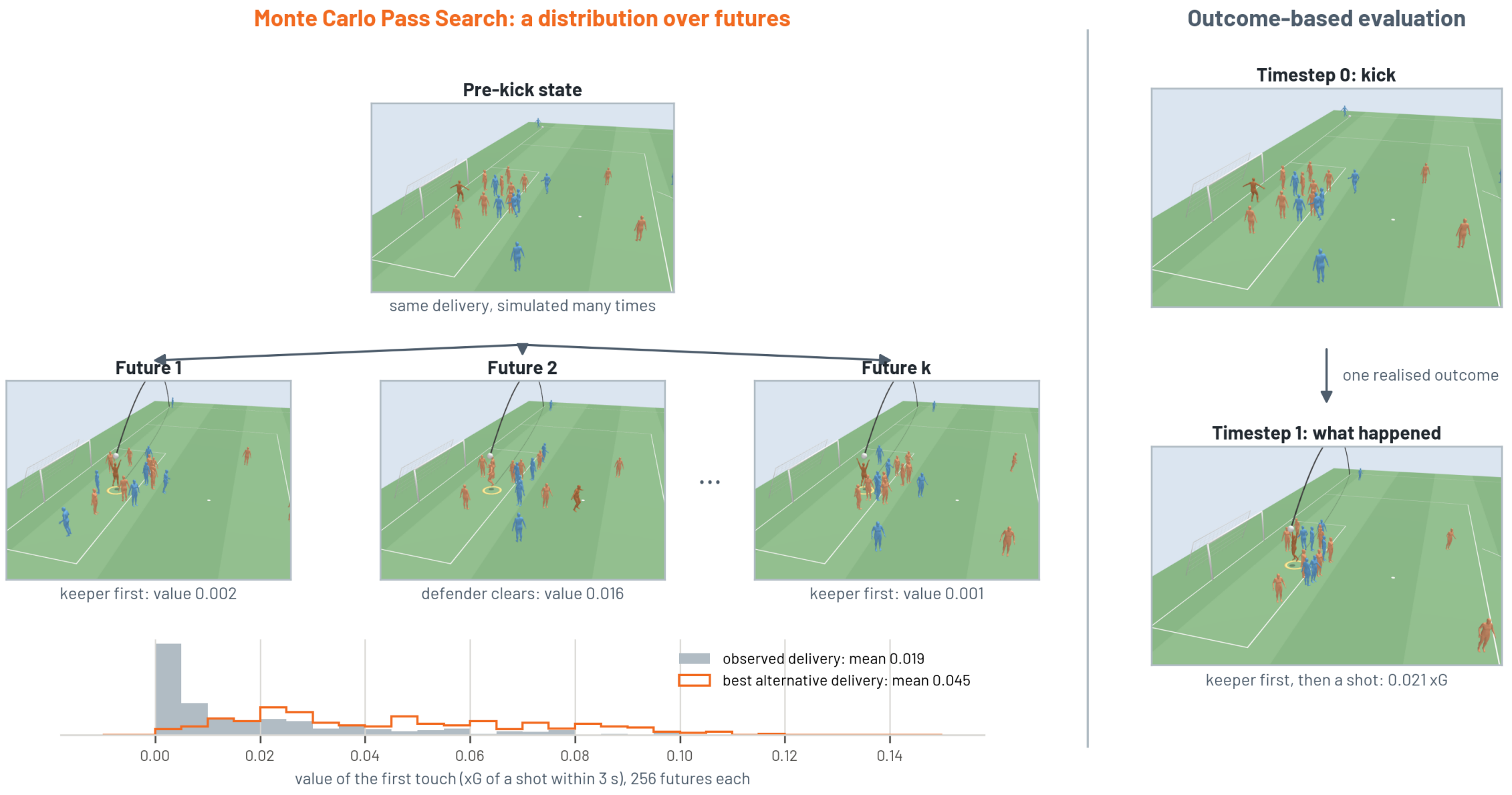

**Figure 1:** From point estimates to distributions. Right: outcome-based evaluation scores the one future that happened (here the keeper reaches a Napoli corner first and the shot that follows is worth 0.021 xG). Left: MCPS plays the same pre-kick state forward many times with a world model, values every future, and compares the observed delivery's value distribution with that of the best alternative.

### Contributions.

- **A set-piece world model built from standard event and tracking data**, including 3D ball flights reconstructed from tracking that never measures the ball's height. Each component is a separate model, validated on held-out matches, so it can be checked, replaced or retrained on its own.
- **Local, global and set-up search.** Local search samples execution variants of the observed delivery (how sensitive was it to the kick?); global search samples alternative deliveries (was a better ball available?); set-up search samples where the attackers stand at the kick, with the defending team's set-up responding through a learned model.
- **An evaluation protocol for decisions made with learned simulators**: fresh futures, an independent value model as the judge, and a realism constraint on recommendations. It changes the answer more than any modelling choice.
- **An evaluation of every set piece, team and player** on all three decisions (ball, set-up, runs), with player rankings built from the same counterfactuals.
- **A data-needs study** for professional use: which components still improve with more seasons, and roughly how many seasons each needs.

## 2 How MCPS works

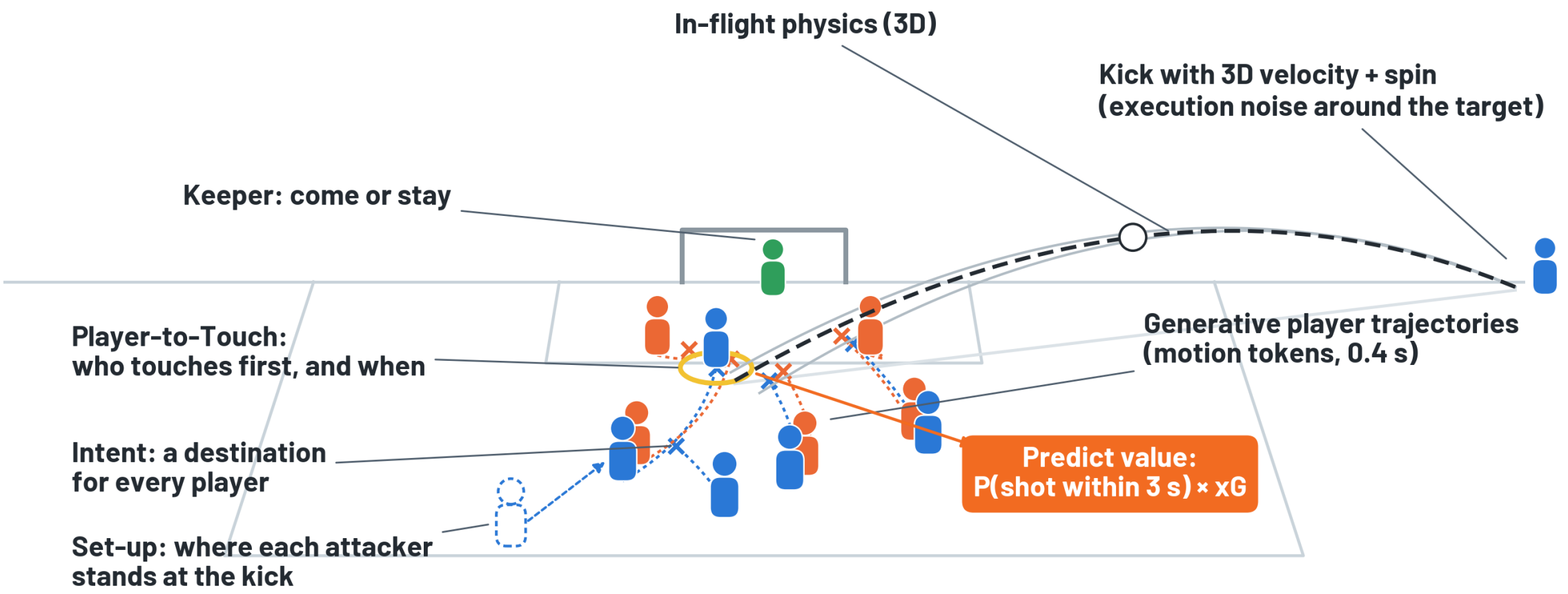

**Figure 2:** Workflow for each simulated set piece. Every labelled step is a separate model; the policy chooses the kick (and, in set-up search, where the attackers stand), the world model plays it out to the first touch, and the value model scores it.

Each future follows the steps in Figure 2.

- **Kick parameters.** A delivery is a 3D launch velocity plus spin. For the observed delivery we infer them by fitting ball physics to the event and tracking data. Alternatives are a target point in the box (8 depths × 13 lateral positions) at one of three heights (apex 3, 4.5 or 7 m): about 180 feasible options per set piece. A delivery is a shot, not a cross, and is never considered if its intended flight would enter the goal untouched or drop onto the goal line (within 1 m of it, between the posts and under the bar), or if at least 1 in 10 of its executions would go in (95 % of real corners stay below that rate); a real delivery that is a shot is kept as the baseline but never recommended. The taker never hits a target exactly, so every kick receives execution noise (5 % speed, 2.5° direction, 2° elevation, spin).
- **In-flight physics.** Gravity, drag, spin and bounce fly the ball. The key output is where and when it comes down through head height (2.3 m), the drop point L.
- **Set-up.** Where every player stands at the kick: the observed positions, or a searched set-up for the attackers with the defending team's response drawn from a learned set-up model (Section 3).
- **Intent.** A destination model draws, for every player, where he is heading when the ball arrives, from the scene at the kick and L [19].
- **Generative player trajectories.** A motion-token transformer in the style of SMART [18] (256 tokens of 0.4 s, 6.7 M parameters) moves each player towards his destination with realistic path spread and body spacing.
- **Keeper.** A come-or-stay model decides whether the keeper attacks the ball; if he comes, he runs to it after a reaction delay.
- **Player-to-Touch.** A survival model gives, every 0.08 s, the probability that each player makes the first touch; the ball may also go untouched.

- **Value model.** The first touch is worth the probability that the attacking team shoots within 3 s, times that shot's xG. Defenders' and keepers' touches keep a small value (second balls).

**Intent, then execution.** A generator that only predicts each player's next 0.4 s from his recent motion is accurate step by step, but loses the purpose of a run within about a second. On held-out set pieces, real players near the drop point close in on it by 0.61 m in the first 1.6 s; such a generator closes in by 0.02 m, because every random step reads as a change of plan. Drawing a destination once per player and executing it keeps the purpose: with the true destinations the generator reproduces the real 0.56 m.

**Calibrated to cause.** Because the destination model sees L, it could simply follow the ball and make simulated players react to every candidate delivery more than real ones do. We measured how far real players move per metre that L moves, holding the pre-kick scene fixed (double machine learning [2] on the part of L the scene does not predict): attackers 0.048, defenders 0.076, keepers 0.058 m per m by the time the ball arrives. One weight per role on the destination model [6] reproduces these responses (0.055 / 0.072 / 0.058).

## 3 Searching deliveries and set-ups

**Two measures.** We keep the terms of the original MCPS [7]. The **surplus** of one option over another is the difference of their values (xG per set piece), each the mean over its simulated futures. The **mean surplus** of what happened is its value minus the mean value of its simulated futures: positive when it turned out better than a typical future of the same choice.

**Local search: execution sensitivity.** Execution variants of the observed delivery show how much its value depended on the kick. The mean surplus of what happened over the observed delivery's futures separates a good decision that went wrong (negative) from a poor decision that got lucky (positive).

**Global search: the best alternative.** Every alternative delivery is screened with 8 futures; the best ten and the best realistic ones are re-tested with 48. We report the **surplus of the best alternative** over the observed delivery, both in the same scene.

**Set-up search: where attackers stand.** The same machinery can move the attackers before the kick. A **set-up model** predicts where each player stands at the kick given the players already placed, one player at a time. Each player is either placed relative to an already-placed player (marking, stacking, screening the keeper) or in a zone relative to the goal; the defending team is placed after the attackers, so its markers follow the attackers wherever they are moved. Trained on 26 k open-play situations and the set pieces of the training matches, its sampled defences mark as tightly as real ones (on held-out matches a box attacker has a defender within 1 m 24.2 % of the time, 24.0 % in tracking). For every set piece we sample 19 alternative attacker set-ups from the model (half move every attacker in the box, half move one to three of them), draw a fresh defending set-up for every future, let every player choose a destination from his new starting point, and search set-up and delivery together.

**Keeping the search honest.** Picking the maximum of noisy estimates flatters the winner, and with learned models in more than one way [15]. Three rules keep the reported gain honest.

- **Fresh futures.** Chosen options are re-valued on new futures, never on those used to choose them.
- **An independent judge.** The search also favours options whose value its own value model happens to overestimate, and fresh futures do not remove that. We search with one 5-seed ensemble [8] and judge with a separately trained one (or with models trained on disjoint halves of the matches). For set-up search the defenders' set-up model is also swapped for an independently trained one.
- **Realistic recommendations.** We also report the best delivery that resembles real ones (its distance to the five nearest real deliveries of other matches within the 75th percentile of real deliveries' own).

## 4 Results

**Data.** Serie A 2025/26 event data (StatsBomb) and 25 Hz tracking for the 30 matches among six clubs; 249 lofted deliveries (170 corners, 79 free kicks into the box). Five match-grouped folds: every number comes from models that never saw that play's match. Everything runs on one GPU (about 9 s per set piece for the delivery search, 20 s with the set-up search).

**Table 1:** Surplus of global search's best delivery over the observed one (249 set pieces; ± s.e. over plays).

| Surplus over the observed delivery (xG per set piece) | Best delivery | Best realistic delivery |
|---|---|---|
| Judged by an independent value ensemble (main result) | 0.0085 ± 0.0006 | 0.0079 ± 0.0007 |
| Judged by models trained on disjoint halves of the matches | 0.0084 ± 0.0007 | 0.0083 ± 0.0007 |
| Judged by the search's own value ensemble | 0.0087 | 0.0081 |
| Corners (170) / free kicks (79), independent judge | 0.0083 / 0.0090 | 0.0082 / 0.0073 |
| Set pieces where the searched delivery is better | 82 % | 74 % |
| Observed delivery's value (independent judge) | 0.0307 xG | |

**What the better ball looks like.** Averaged over set pieces, the best realistic delivery raises the attackers' share of first touches from 0.28 to 0.38 and cuts the keeper's from 0.13 to 0.06; balls nobody touches fall from 0.14 to 0.09. It is flatter (apex 4.3 vs 5.0 m; the 4.5 m profile in 114 plays, 3 m in 97, 7 m in 38), slightly more central (3.9 vs 4.1 m from the middle of the goal) and at about the same depth (6.3 m from the goal line). Per team-season, 0.0085 xG per delivery is about 1.4 xG.

**One set piece in depth.** Figure 3 takes one Napoli corner apart. The keeper reached the real delivery first and the attack's follow-up shot was worth 0.021 xG (a). Global search finds a lower, faster ball (apex 4.5 m, 31 m/s) that drops 7.5 m from the goal line instead of 4.1 m: the real 7 m lob gives the keeper time to come. Over 256 futures each (b), the observed delivery is worth 0.019 xG and the searched one 0.045, a surplus of +0.026; what actually happened has a mean surplus of +0.002 over the observed delivery's futures, so the taker executed his choice about as well as usual and the loss came from the choice. Most executions of the observed delivery come down where the keeper arrives first (keeper first in 49% of futures); the searched ball moves the first touch from the keeper to the attackers (c: attackers 0.16 → 0.42, keeper 0.49 → 0.01). The set-up search (d) moves two attackers towards goal (one from 30 m to 15 m and one from 18 m to 11 m from the goal line), but for this corner the set-up matters little: with sampled defences the observed play is worth 0.024 xG, the best set-up 0.020, the best ball 0.048 and both together 0.046. Here the ball is the decision that matters.

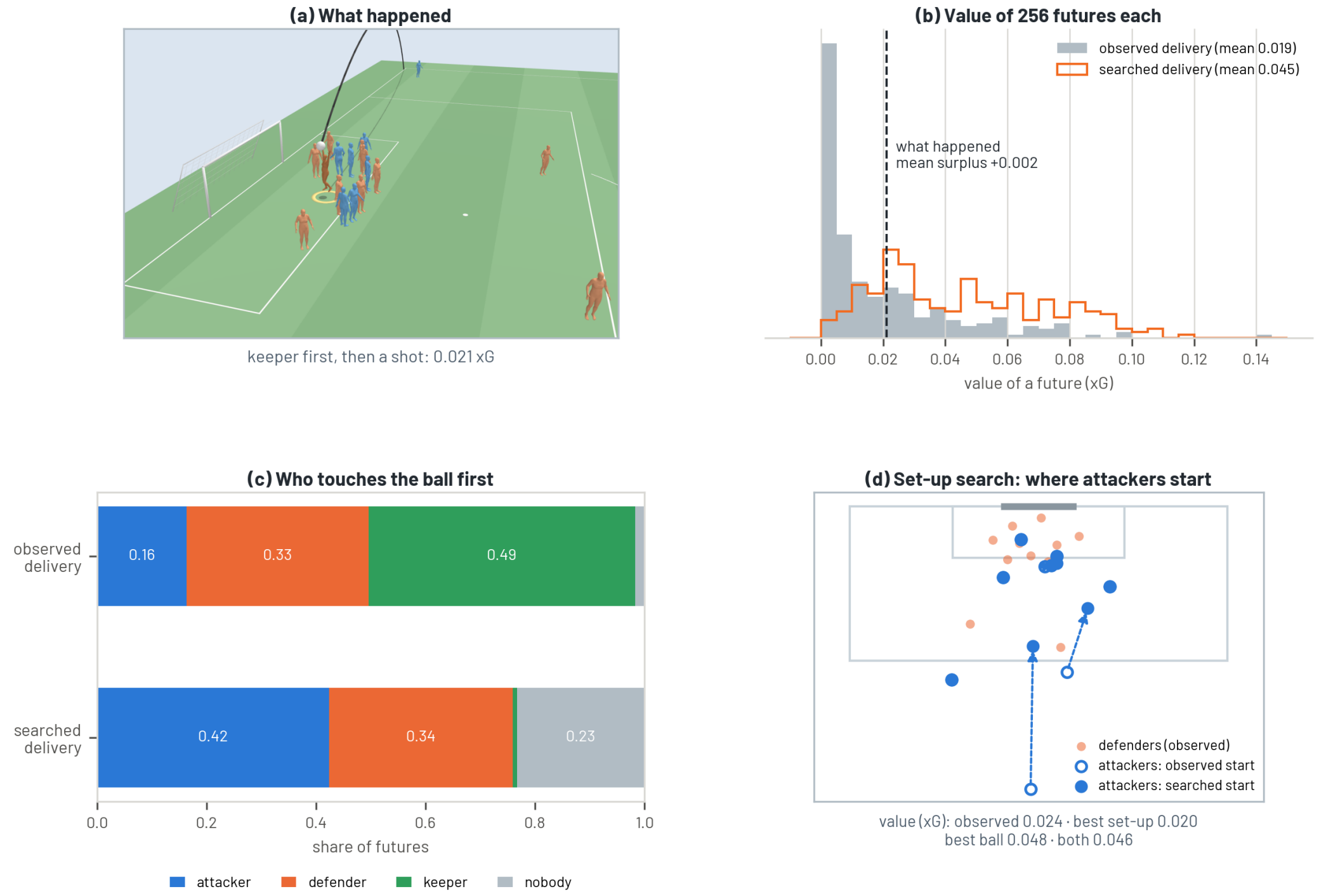

**Figure 3:** Counterfactual analysis of one held-out corner (Napoli, 2025-09-28). (a) What happened: the keeper touches first, then the attack shoots. (b) Value of 256 futures each of the observed and the searched delivery; dashed: what happened. (c) Who touches first. (d) Set-up search: observed (open) and searched (filled) starting positions of the attackers.

**Searching the set-up.** Table 2 compares the set-up search with the delivery search, all with the defending set-up drawn from an independently trained set-up model and valued by an independent value ensemble. With the observed ball, the best of 20 set-ups gains +0.0027 xG over the observed set-up; the best delivery alone gains +0.0074; searching both gains +0.0108. On top of the best ball the set-up adds +0.0033 (corners +0.0037, free kicks +0.0026), more than designing the runs (+0.0012, below). Designing the attackers' runs as well, for the best three set-up and delivery combinations, adds +0.0007 on top of the best set-up and delivery, and searching all three together gains +0.0124 over the observed play (Table 2). The better set-ups mostly bring one more attacker into the box at corners (6.5 → 7.6, 1.1 fewer waiting outside it) and more into the six-yard box (2.2 → 2.9); set-ups sampled at random from the model have 7.4 and 2.3, so the search selects rather than copying the model's habits. The search's own models, the independent ones and a differently trained destination model agree (Appendix E). One caveat matters for coaches: the value model counts only the attacking team's shots within 3 s, so the counter-attack risk of pulling the edge-of-box player forward is not priced.

**Table 2:** Set-up search (independent set-up model and value ensemble as judge; ± s.e. over plays). Every row uses sampled defenders. The last two rows also design the attackers' runs (2.8 m destination zones) for the best three set-up and delivery combinations and are judged on separate fresh futures.

| Gain over the observed set-up and delivery (xG per set piece) | All (249) | Corners | Free kicks |
|---|---|---|---|
| Best delivery, observed set-up | +0.0074 ± 0.0007 | +0.0073 ± 0.0008 | +0.0078 ± 0.0010 |
| Best set-up, observed delivery | +0.0027 ± 0.0003 | +0.0027 ± 0.0004 | +0.0026 ± 0.0006 |
| Best set-up and delivery together | +0.0108 ± 0.0007 | +0.0109 ± 0.0009 | +0.0105 ± 0.0010 |
| … realistic deliveries only | +0.0096 ± 0.0007 | | |
| Set-up on top of the best delivery | +0.0033 ± 0.0004 | +0.0037 ± 0.0005 | +0.0026 ± 0.0005 |
| Set-up, runs and delivery together | +0.0124 ± 0.0007 | +0.0120 ± 0.0008 | +0.0134 ± 0.0011 |
| Runs on top of the best set-up and delivery | +0.0007 ± 0.0002 | +0.0002 ± 0.0002 | +0.0018 ± 0.0005 |

**Robustness.** The best delivery's surplus survives every alternative we tried (Appendix D): five independently trained pipelines (the best delivery is better for a majority of them in 70 % of plays), players reacting to the ball at the observational upper bound (−0.0001), a fixed keeper rule (+0.0004), four times more ball spin (−0.0006), defenders that respond to the attackers' destinations (−0.0016) and soft collision handling (+0.0002). Individual recommendations are less stable: independently trained pipelines scatter the best target by a median 2.7 m, so they should be read as zones, not points.

**Designing the runs.** Searching the attackers' run destinations (a 2.8 m zone per attacker) together with the ball adds only 0.0012 xG over the best ball (free kicks 0.0015, corners 0.0010), and a fixed routine, the three nearest attackers attacking the drop zone, adds 0.0023. With the set-up searched as well, designed runs add +0.0007 on top of the best set-up and delivery. Runs are worth little because there is little time: between the kick and the ball's descent real attackers move a median 2.2 m (90 % less than 5.5 m), so where an attacker stands at the kick decides most of where he meets the ball, which is why the set-up matters more. Designed runs do take effect, and simulated runners gain as much separation from defenders as real ones (0.18 vs 0.13 m per metre run).

**Every set piece, every team, every player.** The same machinery evaluates each of the 249 observed set pieces on all three decisions (Figure 4, top). Per set piece, the surplus of a better ball is +0.0085 xG, of a better set-up +0.0027 and of designed runs +0.0011. The observed set-up has a mean surplus of +0.0013 over the set-ups the model proposes for the same players, and is worth less than their mean in 37 % of set pieces. The exact kick has a mean surplus of +0.0024 over noisy executions of the same delivery, and what happened one of +0.0019 ± 0.0015 over its simulated futures, close to zero as it should be if the futures are calibrated. Teams differ mostly in the ball: Juventus and Napoli left 0.0116–0.0125 xG per set piece with it, Inter 0.0066. Real players move better than the model's typical players on both sides (attackers' runs +0.011 xG per set piece, defending 0.009), so the two largely cancel; making simulated players less passive is the clearest use for more data. Ranking players by the same quantities (Figure 4, bottom), Baturina, Dybala, Dimarco leave the least on the table per delivery (0.0057–0.0059 xG) and Yildiz the most (0.0144);

Pavlović, Thuram, Gatti gain the most from their real movement relative to the model, and Ndicka, Akanji, Butez take the most away. With at most ten matches per player these rankings are descriptive. The largest single examples, real and searched, are listed in Appendix D.

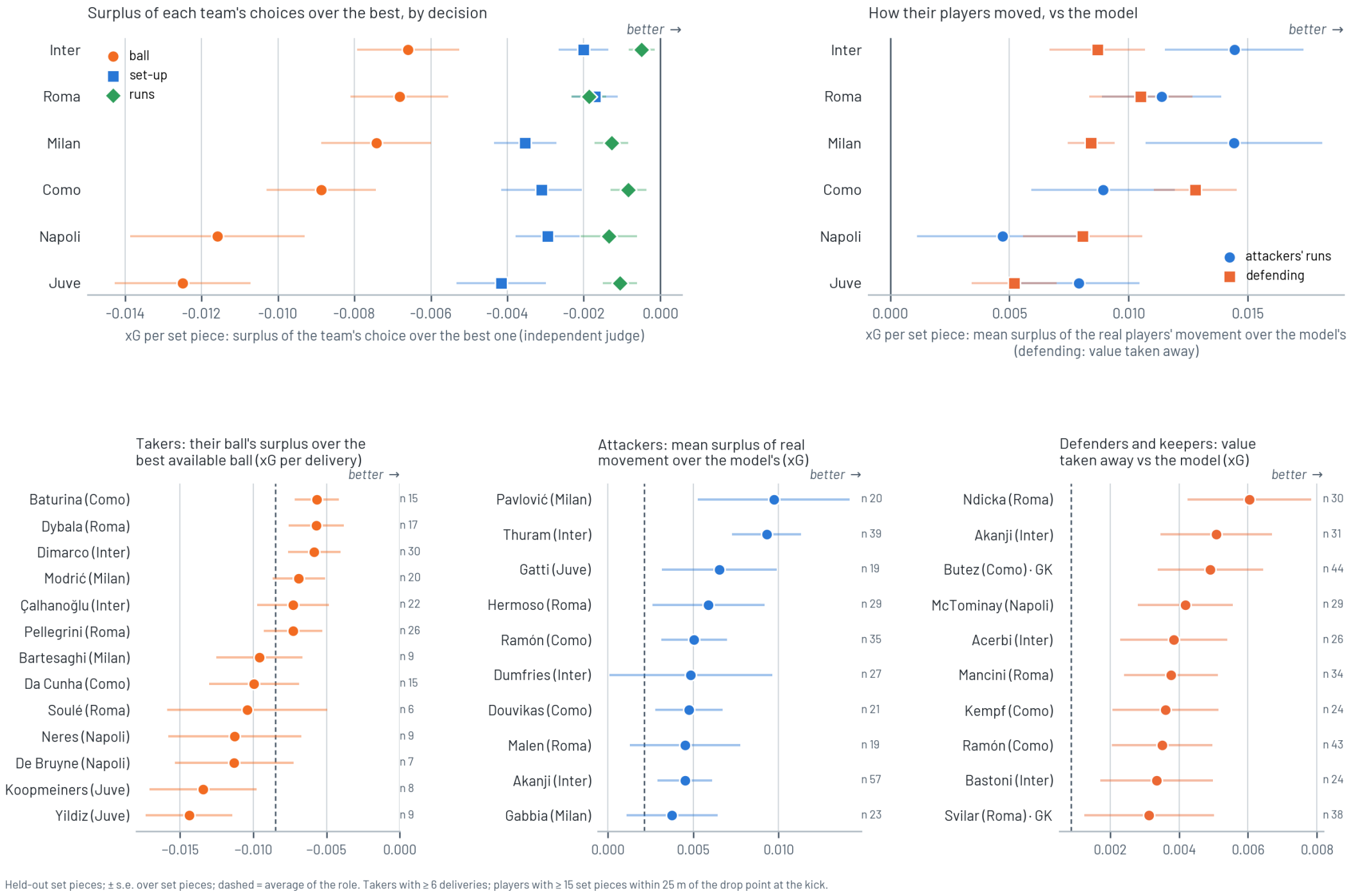

**Figure 4:** Evaluation of every observed set piece (held-out models, independent judges). In every panel higher is better. Top left: surplus of each team's ball, set-up and runs over the best choice the search found (xG per set piece; 0 = nothing better existed). Top right: mean surplus of the real players' movement over the model's from the same start, for each team's attackers and defenders. Bottom: takers by their ball's surplus over the best available one; attackers by the mean surplus of their real movement; defenders and keepers by the value their real movement took away (± s.e. over set pieces; dashed: role average).

## 5 What is new

Most football decision models value the action that happened. xT, VAEP and EPV value states or actions [3, 4, 14]; expected-pass models [1, 10] and the physics-based pass model of Spearman et al. [16] estimate a pass's success, and the latter can probe hypothetical passes under hand-set assumptions about how players move. TacticAI [17] predicts the receiver and shot of a corner from a snapshot at the kick and suggests position adjustments. Set-piece studies otherwise describe outcomes [11] or cluster routines [13]. MCPS [7] searched pass variants with a generative world model in open play. MCPS for set pieces adds:

- **Counterfactuals of the delivery and the set-up, resolved in time.** For every candidate we simulate the 3D flight with execution noise and the response of all 22 players, so its value reflects run timing, marking, the keeper's decision and second balls, rather than a snapshot.
- **A world model calibrated to cause.** Separating intent from execution keeps simulated runs purposeful, and matching the players' response to the ball to a causal estimate stops the simulator from over-reacting to counterfactual deliveries. We know of no sports simulator with this identification step.

- **A decision-grade evaluation protocol.** Independent judging of the value model, standard in causal inference but new to sports decision analysis as far as we know, removes most of the apparent gain of a naive search (0.014 → 0.004 with a single value model).
- **3D from data that never measures the ball's height**, by fitting physics to the kick, the contact event, the toucher's tracked position and his height (held-out contact-height error 0.10 m).

## 6 From 30 matches to professional use

A club cannot act on a model fitted to 30 matches of six teams; it needs one trained on its league over several seasons and refreshed as squads change. MCPS is built for that. Each component is trained, validated and replaced on its own, against its own held-out check (Appendix B), while the search and the evaluation protocol stay fixed, so a better or bigger component drops in without re-validating the whole system. At our data size, retraining every component for one fold takes about 40 minutes on one GPU (several hours per season of data), and searching a full season (about 3,150 deliveries) about 8 GPU-hours for deliveries and about 18 for deliveries and set-ups.

**Two kinds of data.** Open-play data is cheap: we already pretrain on 26 k open-play situations from 30 matches, and a season holds about 330 k. Set-piece data is scarce: about 150 training set pieces per fold today, about 3,150 per season. The motion components lean on the cheap kind: without open-play pretraining the simulated paths are 9 cm worse on average, and pretraining on lofted passes only (8× less data) costs 3 cm. The Player-to-Touch and keeper models gained nothing significant from open-play pretraining, and the value model already pools all lofted contacts; these three need more set pieces.

**How much more?** We retrained each component on a random 1/4, 1/2, 3/4 and all of its training matches (several draws each, all five folds), scored it on the held-out matches, and measured the gain from each doubling of the data and whether that gain is shrinking (Figure 5, Appendix G) [5]. The **Player-to-Touch model** improves fastest (20 % of its loss per doubling) but is clearly saturating: each doubling buys about 55 % less than the one before, so about one season captures most of what more data can give (median 0.7 seasons; the upper end of the interval is long). The **keeper model** (12 % per doubling) and the **value model** (6 %) show no saturation yet: keeper claims and set-piece shots are rare events (30 keeper first touches and 60 attacker shots in our data). Even if their gains started halving now, they would need about 1 and 0.5 seasons; at their current pace, several seasons keep paying off, so they need the most new set-piece data. The **destination** and **motion** models gain 1.7 % and 0.8 % per doubling of set-piece data; for them, more open-play data and better inputs matter more than more set pieces.

**What more data will not fix, and what it will.** More seasons of tracking will not reveal a team's planned runs, which are invisible at the kick; a club can supply them (its own routines as the attackers' intent), the one input we cannot learn. What data will fix is the noise that makes single recommendations unstable: recommendations of independently trained pipelines scatter by 2.7 m today, and if that spread is estimation noise it shrinks with the square root of the data, to about 1 m with one season (an extrapolation). With several seasons, team and taker identity can enter the destination, set-up and keeper models, turning league-level advice into opponent-specific advice.

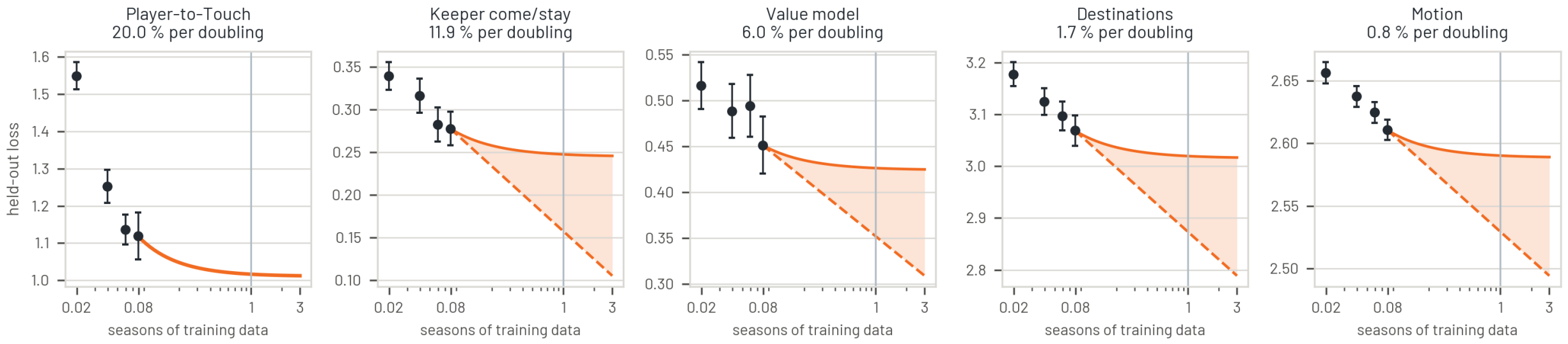

**Figure 5:** Learning curves: held-out loss of each component against seasons of training data (log scale; we have 0.08). Points: mean over folds and draws (± s.e./2). Player-to-Touch, whose gain per doubling is measurably shrinking: projected curve. Others: band between the gain halving each doubling (solid) and the current trend continuing (dashed).

## 7 Limitations and conclusion

Only direct deliveries are searched: the ball lofted into the box from the corner or free-kick spot. 52 corners and 199 attacking-half free kicks were played short or low (routines with a pass before the cross) and are not examined, although the possessions they start carry 28 % of set-piece xG (0.052 xG per short corner vs 0.055 per direct one). The simulation has no fouls (holding, blocking, penalties) and no individual strengths or weaknesses: players differ only by role, height and position, not by heading, jumping or keeping ability. There is no body contact, aerial duel, header physics or collision: bodies are only kept apart, and who touches the ball first is a probability.

Thirty matches among six strong clubs bound everything here: per-club and per-player numbers are descriptive, and the world model has seen few unusual routines. After the kick the simulated defence marks the drop zone less tightly than real defences (33–37 % of attackers near it unmarked vs 24.5 % in tracking), so simulated attackers win slightly too easily away from the planned target. The keeper is under-predicted for balls dropping 3–6 m out. Attackers' planned runs are not visible at the kick, so the world model draws them from what similar scenes did; a club that supplies its own routine closes that gap. Play after the first touch is valued, not simulated, and only for the attacking team's next 3 s: counter-attack risk, which matters most for set-up advice, is not priced.

Within these limits the message is consistent across every check: for most set pieces a better delivery was available, worth about 0.0085 xG (≈1.4 xG per team-season), and it is the one that gets the ball to an attacker before the keeper. The same machinery rehearses set-ups: choosing where the attackers stand adds about 0.0033 xG on top of the best ball, mostly by putting one more attacker into the box, while designed runs add little (+0.0007 on top of the best set-up and ball) because players move only about 2 m while the ball is in the air. The practical lesson for anyone building such tools: a search over a learned simulator flatters itself in ways held-out accuracy does not reveal, and must be graded by models that did not make the choice.

# Appendix

All numbers are from held-out folds.

## A. Data, selection and 3D ball reconstruction

- **Data.** 30 Serie A 2025/26 matches (double round robin of Milan, Inter, Juventus, Napoli, Roma, Como): StatsBomb events and 25 Hz tracking. Tracking has jersey numbers only, the ball's height is always 0, and the ball is interpolated during lofted flights. Event and tracking clocks are aligned per period by matching the ball to pass and restart events (60/60 periods, median 2.4 m). Players are mapped to tracking columns (925/926 player-matches), and heights come from public sources (141/141 players).
- **Selection.** Corners and free kicks with a high pass; free kicks only from the attacking half into the box: 170 + 79 = 249 deliveries. The first contact comes from the event feed (attackers win it in 35 % of corners, 40 % of free kicks).
- **Ball.** RK4 flight with quadratic drag ($k = 0.0133\ m^{-1}$), spin $a = c\,\hat{z} \times v$ and bounce. A MAP fit of (end point, contact height, flight time, spin) uses the event end ($\sigma$ 1.5 m), the tracked toucher at contact ($\sigma$ 1.0 m, 93 % of plays), a contact-height prior (header = player height + 0.40 ± 0.13 m, from 128 headed shots), the event contact time ($\sigma$ 0.12 s) and the spin sign from the in/out-swing label (0.06 ± 0.04 $s^{-1}$). 99 % of arcs are physically plausible (median 27 m/s, 22°, apex 4.4 m, 1.6 s). Held-out contact-height error is 0.104 m (a constant prior gives 0.113 m), and the real toucher is a median 0.79 m from the reconstructed ball at contact. Propagating the fit's posterior changes a single kick's value by about 0.006 xG and no aggregate result.

- **Situations.** 76 frames at 12.5 Hz (2 s before and 4 s after the kick), 22 slots (11 attacking, 11 defending, goalkeepers first), always attacking +x. After the kick the ball follows the untouched reconstructed arc, in training exactly as in the search, so no model sees where the real contact happened.

## B. Components

**Table 3:** The components. Each is trained on the training folds only; a validation fold is used for early stopping.

| Component | Model | Trained on (per test fold) | Selection / calibration |
|---|---|---|---|
| Set-up | 4-layer transformer over 22 players + scene token; per player a mixture of a zone grid (1 m) and offsets (0.5 m bins, ±5 m) from each already-placed player; players placed one at a time | 26 k open-play situations, then ~150 set pieces | best validation-fold likelihood; sampling temperature 0.65 matched to real marking rates |
| Destinations (intent) | 4-layer transformer over 22 players + scene token; per player a 40×40 categorical (0.7 m bins, ±14 m) of the displacement to the 0.4 s step nearest the drop time | 26 k open-play situations (lofted ×4), then ~150 set pieces | best validation-fold NLL; per-role weights −0.6 / −0.35 / 0 (att / def / GK) matched to the causal response |
| Motion (execution) | space-time transformer (d 256, 6 blocks, 6.7 M), 256 motion tokens of 5 × 0.08 s displacements (k-means on 3 M open-play tokens); trained with the real destinations as input (dropped 20 %) | 26 k open-play situations, then set pieces | best validation-fold token CE; nucleus 0.95; soft repulsion between players (λ 3, 0.45 m) |
| Keeper | logistic come/stay on 16 kick-time features (keeper–L distance, hang time, required speed, crowding); if come, runs to L after a delay | 237 set-piece aerial balls | L2 and delay chosen out of fold on first-toucher likelihood |
| Player-to-Touch | discrete-time survival model, hazard + softmax over 21 players, 3-layer transformer (d 128); ball height, reach margin, drop point | set pieces (3 seeds) + lofted-pretrained (3 seeds) | 6-model ensemble |
| Value model | two-head MLP: P(attacking shot within 3 s) and xG of that shot, from 22 contact-instant features and the toucher's role | 5,406 lofted first contacts (set pieces + open play) | 5-seed ensemble; judged by a separately trained 5-seed ensemble |

## C. World-model validation

**Table 4:** Held-out validation of the components (details below and in Appendix E).

| Is the world model trustworthy? | Simulated | Reference |
|---|---|---|
| Accuracy of simulated futures vs a next-step token generator (energy score) | −5 to −7 % (all roles) | lower is better |
| Response to the ball (m per m of L): att / def / keeper | 0.055 / 0.072 / 0.058 | causal estimate 0.048 / 0.076 / 0.058 |
| Who touches first, real movement: NLL / top-1 / attacker AUC | 0.995 / 0.70 / 0.91 | 21 candidates |
| Total value of the 249 observed deliveries | 7.7–8.2 xG | 8.59 xG realised |
| Keeper come/stay (set pieces) | AUC 0.93 | beats a calibrated rule |
| Players closer than 0.3 m (share of player-frames) | 1.78 % | 1.79 % in tracking |
| Sampled defences: box attackers with a defender within 1 m / 1.5 m | 24.2 % / 43.4 % | 24.0 % / 44.5 % |

**Accuracy and commitment.** Energy score of 32 simulated futures against the real path (kick to drop, per role) improves on a next-step token generator by 0.034 / 0.031 / 0.031 m for attackers / defenders / keepers (5.8–6.6 %).

The spread-to-error ratio is 1.00–1.02: the futures are neither over- nor under-dispersed. Convergence on L (players within 6 m, first 1.6 s): real 0.61 m; next-step generator 0.02 m (0.35 m when fed the real history, 0.23 m when it moves by expected rather than sampled steps, showing that sampling noise erases intent); with destinations 0.56 m.

**Causal response (double ML).** Treatment L; nuisance models for E[ L | scene ] and E[ displacement | scene, player ] are cross-fitted gradient boosting; the scene predicts little of L ($R^2$ 0.20 in depth, ≈ 0 laterally). Response by time after the kick (attackers / defenders): 0.16 s 0.001 / 0.000 (placebo), 0.8 s 0.013 / 0.016, 1.2 s 0.033 / 0.041, at the drop 0.048 ± 0.024 / 0.076 ± 0.016 (keeper 0.058 ± 0.016). Applied to the world model's own rollouts, the estimator recovers its known response (0.044 vs 0.050, 0.052 vs 0.046).

**Keeper and first touch.** Out-of-fold first-toucher NLL with simulated players: no keeper model 2.077, calibrated rule 2.044, learned come/stay 1.958 (+0.085 ± 0.044 nats per play over the rule). Player-to-Touch with real movement: NLL 0.995, top-1 0.70, attacker-first AUC 0.91. Ball height is its most important input (+0.6–0.8 nats without it); player height adds nothing once the arc is known.

## D. Evaluation protocol and robustness

**Table 5:** Independent judging of the value model: surplus of each search's best delivery over the observed one (xG), valued on the same fresh futures by every value model. Diagonal = the search grading itself. A single model flatters itself 3- to 5-fold; a 5-seed ensemble does not (+0.0002 for a seed-partner ensemble, +0.0019 for half-data models).

| Search ↓ / judge → | single model | ensemble 1 | ensemble 2 | half 1 | half 2 |
|---|---|---|---|---|---|
| single value model | **0.0143** | 0.0040 | 0.0041 | 0.0040 | 0.0027 |
| ensemble 1 (5 seeds) | 0.0078 | **0.0086** | 0.0084 | 0.0093 | 0.0075 |
| ensemble 2 (5 seeds) | 0.0082 | 0.0086 | **0.0087** | 0.0090 | 0.0073 |
| half 1 (half the matches) | 0.0059 | 0.0069 | 0.0065 | **0.0108** | 0.0067 |
| half 2 (other half) | 0.0083 | 0.0081 | 0.0080 | 0.0096 | **0.0092** |

**Table 6:** Robustness of the best delivery's surplus (paired differences, ± s.e.). Shots at goal are excluded from every search.

| Alternative | Effect on the surplus (xG) | Note |
|---|---|---|
| 5 independently trained pipelines (+ arc uncertainty) | 0.007–0.012 per pipeline | majority positive in 70 % of plays |
| Players react to the ball at the observational upper bound | −0.0001 ± 0.0006 | optima move a median 2.5 m |
| Fixed keeper rule instead of the learned model | +0.0004 ± 0.0007 | optima 7.2 → 6.5 m from the line |
| 4× more ball spin, foot-signed free kicks | −0.0006 ± 0.0008 | picks change worth ≈ 0.001 |
| Defenders respond to the attackers' destinations | −0.0016 ± 0.0005 | same picks re-judged, per-play correlation 0.88 |
| Soft collision handling between players | +0.0002 ± 0.0003 | 41 % same realistic picks, same value |

**Table 7:** Largest single examples (judged xG per set piece; surpluses over the observed choice, or of the real choice over the search's best). Mean surplus: the value of what happened (real kick, real players) minus the mean of the kick's simulated futures. Stills of these and further examples are in Appendix H.

| | Set piece (taker) | xG |
|---|---|---|
| Search: largest gain, ball | Como v Napoli, 46', corner (Da Cunha) | +0.044 |
| Search: largest gain, set-up | Juventus v Como, 94', corner (Conceição) | +0.020 |
| Search: largest gain, runs | Napoli v Roma, 15', corner (Elmas) | +0.018 |
| Search: largest gain, all three | Inter v Napoli, 88', corner (Dimarco) | +0.049 |
| Real ball beats the search's best | Inter v Juventus, 76', corner (Dimarco) | +0.011 |
| Real set-up beats the model's best | Como v Milan, 50', corner (Baturina) | +0.013 |
| What happened: highest mean surplus | Inter v Como, 58', free kick (Çalhanoğlu) | +0.119 |
| What happened: lowest mean surplus | Roma v Inter, 20', free kick (Pellegrini) | −0.036 |

**Evaluating every set piece and player.** Per set piece, in xG, from held-out models. Ball: the surplus of the best (and best realistic) delivery over the observed one, judged; execution = the mean surplus of the exact kick over noisy executions of the same delivery; what happened = the value model on the real kick and the players' tracked movement, compared with the delivery's futures by its mean surplus. Set-up: the surplus of the best of the 19 model set-ups over the observed one, observed ball (judged); the observed set-up's mean surplus over the 19 uses their screening values. Runs: surplus of designed over natural runs, observed ball (judged). Real against model: mean surplus of one team's tracked movement, the other team simulated, over all simulated (48 futures); a player's movement value: only he on his tracked path (24 futures), for players within 25 m of the drop point at the kick. Rankings need six deliveries (takers) or 15 set pieces (players).

## E. Set-up model and set-up search

**Model.** Inputs per player: role, height, taker flag, whether he is already placed and, if so, his position and his velocity over the last 0.8 s; a scene token carries the kick spot and the set-piece type. No ball-flight input: the set-up precedes the delivery. Players more than 35 m from goal, the taker and the defending keeper stay where they are. Training samples a random placed set (attacking team before the defending team, random order within each) and maximises the likelihood of the others; open-play pretraining then set-piece fine-tuning, two independently pretrained families of three seeds each (one searches, the other judges). Knowing the attackers' positions lowers the held-out NLL of each defender by 0.3 nats. The sampling temperature (0.65) was chosen on the validation folds; on the held-out test folds its sampled defences mark like real ones (a defender within 1 m of a box attacker 24.2 vs 24.0 %, within 1.5 m 43.4 vs 44.5 %, six-yard-box occupancy 5.0 vs 5.0; double-marked attackers 15 vs 11 %).

**Search.** A moved player keeps his own pre-kick motion (his track is translated). Free-kick attackers offside at the kick win nothing. Per set piece: the delivery shortlist (best two, best realistic, observed); 20 attacker set-ups (observed + 19 sampled) × shortlist with 24 futures each, executions shared across set-ups; the best three set-ups per delivery and the observed one re-tested on 48 fresh futures; picks judged on 64 fresh futures under the independent set-up family, two destination models and the independent value ensembles. Runs (Section 4): each attacker's destination restricted to a 2.8 m zone, effects estimated by a randomised experiment inside the world model with empirical-Bayes shrinkage. Set-up, runs and delivery together: for the best three set-up and delivery combinations (and the best one with a realistic delivery), runs are designed in the same way starting from the searched set-up (zone probabilities averaged over 32 sampled defences); every combination is then re-valued with natural and with designed runs on 48 paired fresh futures, and the picks are judged as above. Designed runs replace natural ones only where they beat them.

**Further results.** The searched set-up differs from the observed one in 85 % of set pieces, moving the attackers in the box a median 7.2 m. The joint gain is +0.0109 ± 0.0007 under the search's own models, +0.0108 ± 0.0007 under the independent ones and +0.0106 ± 0.0007 with the any-order destination model. Sampled defences are slightly kinder to the attack than real ones (+0.0016 ± 0.0003 xG), so set-up gains are always measured against the observed set-up with sampled defenders. The three-way pick beats the set-up and delivery pick by +0.0015 ± 0.0003 on the same futures, +0.0007 of it from the runs and the rest from re-testing the best combinations.

## F. Collisions and 3D renders

Soft repulsion in the motion-token probabilities (penalty $\lambda \Sigma (1 - d/d_0)_+^2$, $\lambda$ 3, $d_0$ 0.45 m) brings players closer than 0.3 m from 2.37 % to 1.78 % of player-frames (1.79 % in tracking) at no cost in accuracy or decision value. Renders (pyrender, SMPL bodies [9] scaled to each player's height) stage the contact: the Player-to-Touch model's most likely toucher meets the ball at its most likely contact frame, and the ball leaves the contact on an illustrative path.

## G. Learning curves

Each component was retrained with its production configuration on 1/4, 1/2 and 3/4 of the 18 training matches per fold (several draws and seeds) and scored on the same held-out matches (5–30 runs per fraction and component). A power law with an asymptote is not identified by four noisy points, so we report the gain per doubling (slope of the held-out loss over log2(data), relative to the current loss), the ratio $\rho$ of the last doubling's gain to the one before ($\rho < 1$: shrinking) and, where $\rho < 1$ in at least 75 % of bootstrap samples, the data at which a further doubling gains less than 1 % of the loss; otherwise a lower bound that assumes the gain halves from now on.

## H. 3D examples

The examples of the talk's animations, as stills of their last frame. Left: what happened (real players, real ball). Right: one simulated future of the searched choice, the one closest to the median value among futures whose ball passes the shot check (no ball shown is aimed at the goal). Values are judged by the independent models; orange: attackers, blue: defenders.

**Searched balls (observed set-up, natural runs)**

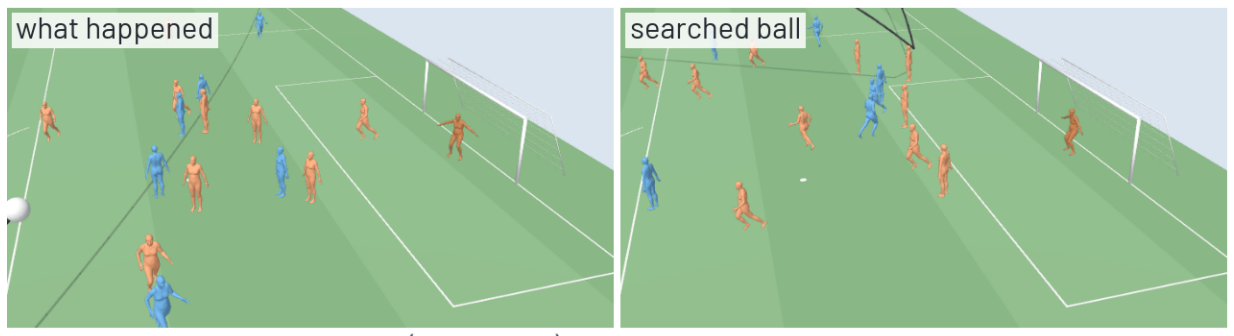

Juventus v Roma, 87', corner (Cambiaso): searched ball +0.038 xG

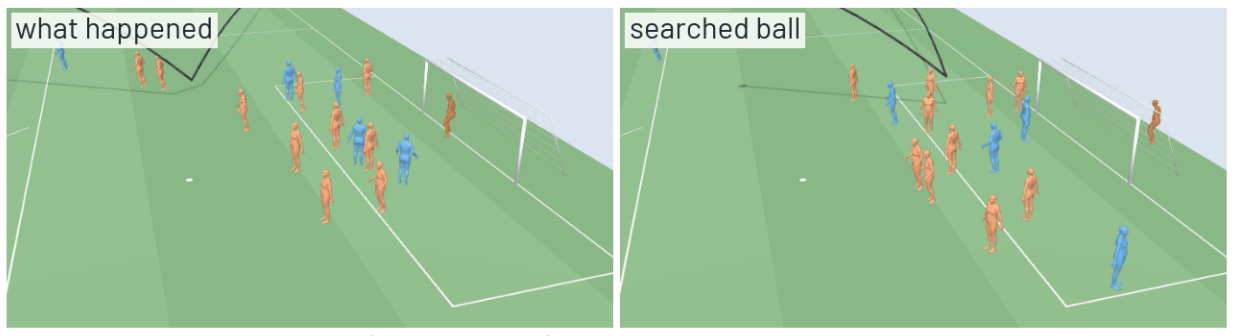

Napoli v Roma, 89', corner (Nascimento): searched ball +0.036 xG

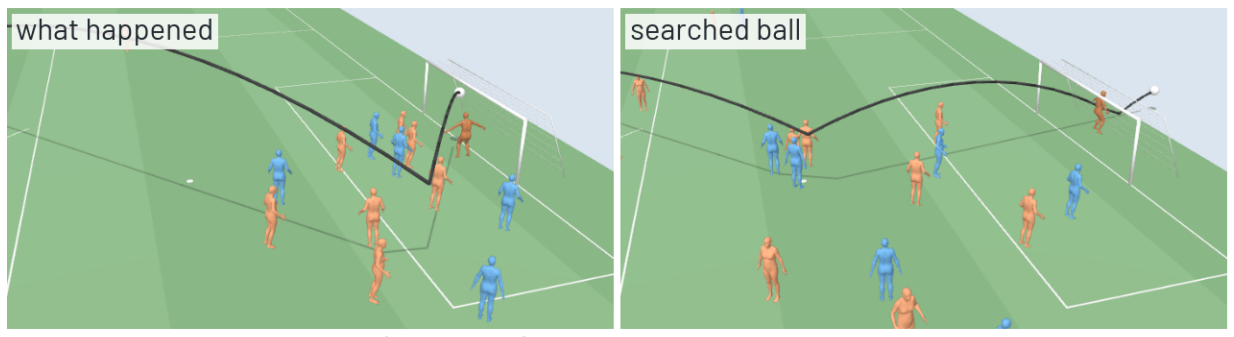

Roma v Como, 17', free kick (Pellegrini): searched ball +0.030 xG

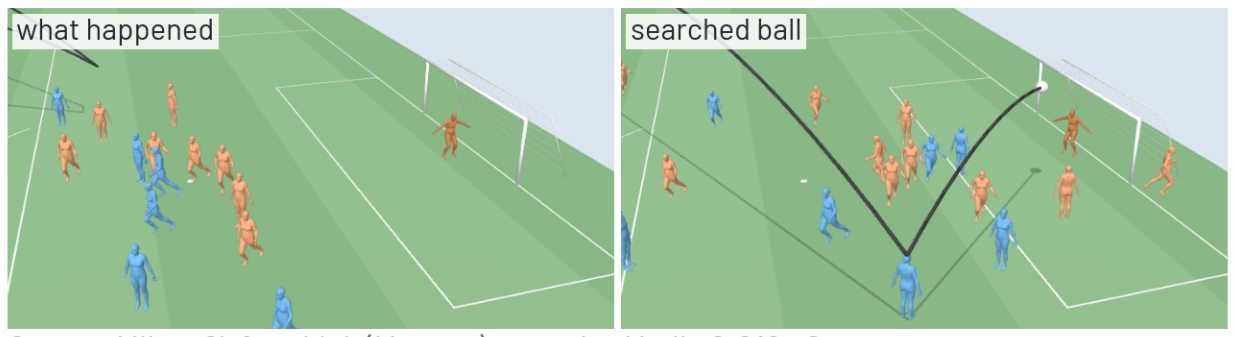

Como v Milan, 2', free kick (Moreno): searched ball +0.019 xG

**Searched set-ups (real ball; defenders placed by the set-up model)**

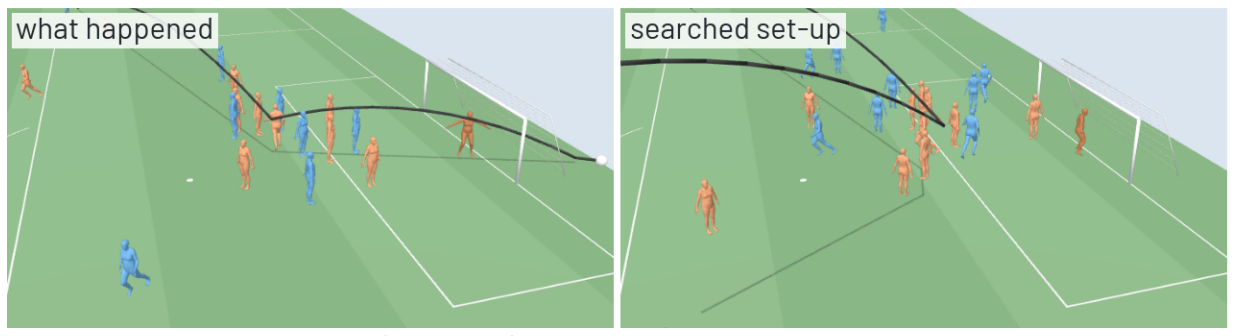

Como v Milan, 89', free kick (Baturina): searched set-up +0.018 xG

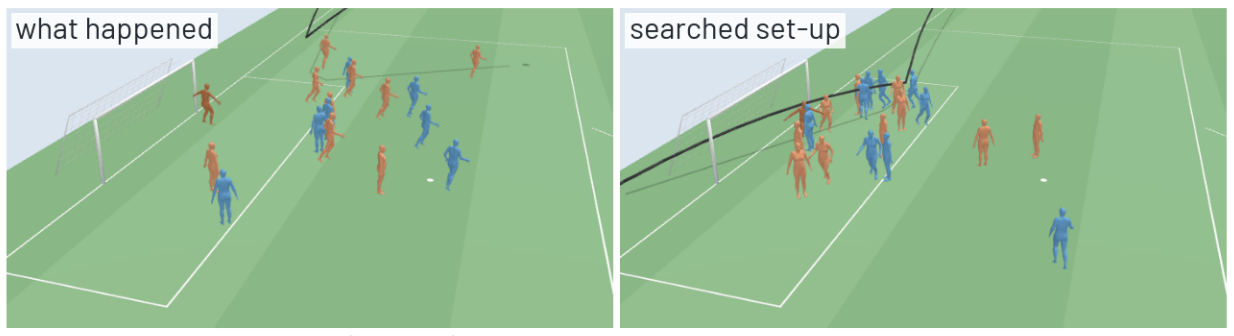

Milan v Como, 51', corner (Modrić): searched set-up +0.017 xG

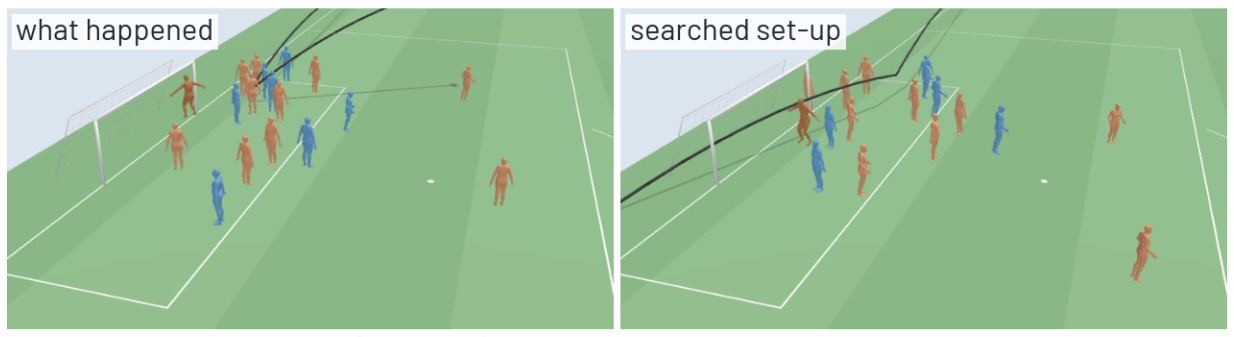

Inter v Napoli, 45', corner (Dimarco): searched set-up +0.013 xG

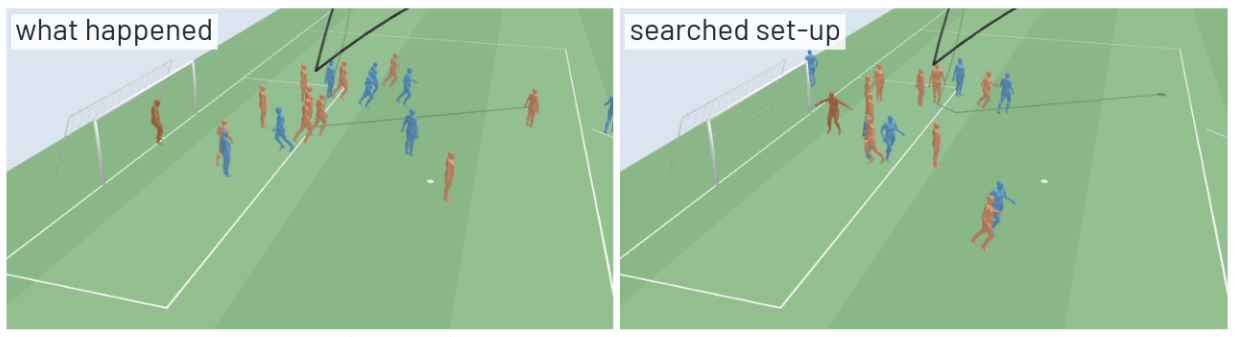

Roma v Milan, 19', corner (Dybala): searched set-up +0.013 xG

**Designed runs (real set-up and ball; discs = target zones)**

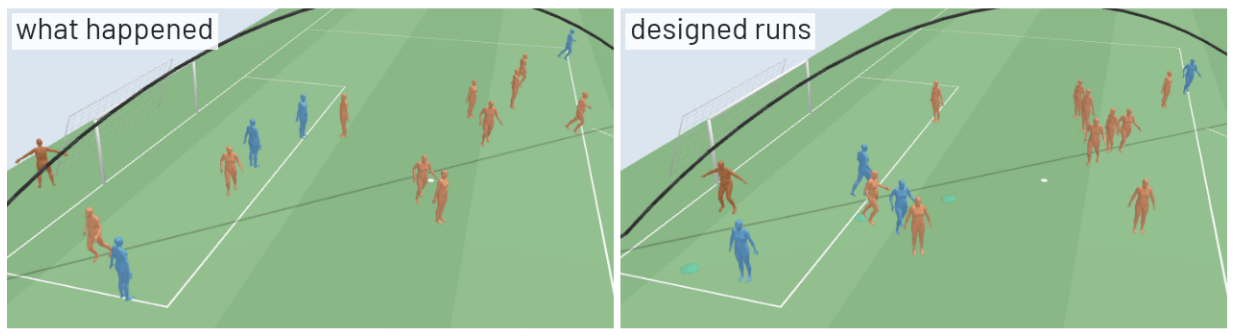

Roma v Como, 75', free kick (Tsimikas): designed runs +0.013 xG

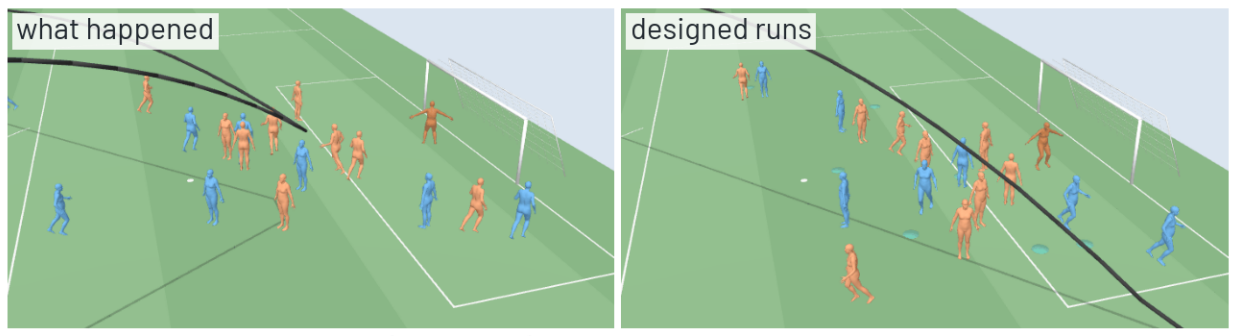

Como v Roma, 57', free kick (Da Cunha): designed runs +0.011 xG

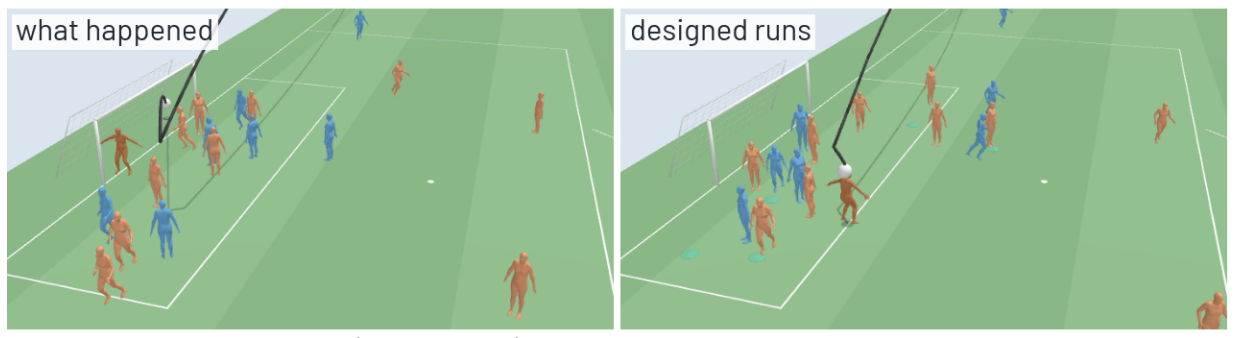

Milan v Roma, 20', corner (Bartesaghi): designed runs +0.009 xG

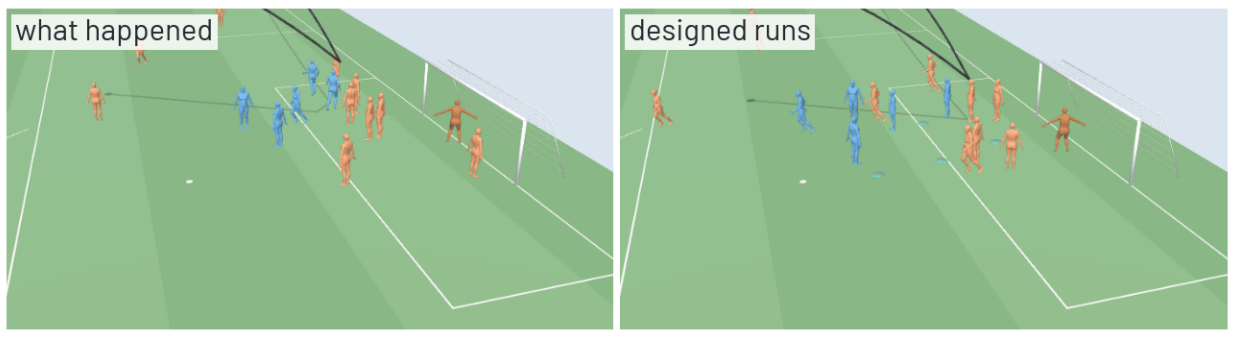

Juventus v Milan, 11', corner (Boga): designed runs +0.009 xG

**Figure 6:** Searched balls, set-ups and designed runs against what happened (large surpluses, one set piece per team where possible).

**Set-up, runs and ball searched together (discs = designed run targets)**

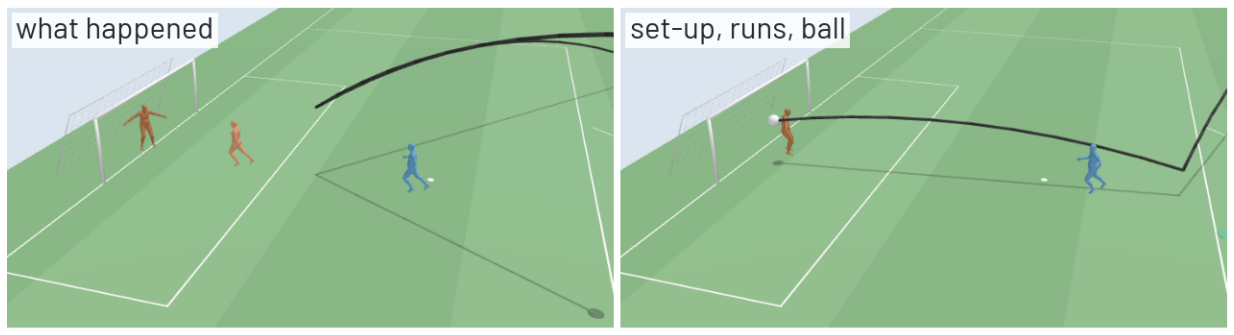

Napoli v Milan, 83', free kick (Elmas): set-up, runs and ball +0.046 xG

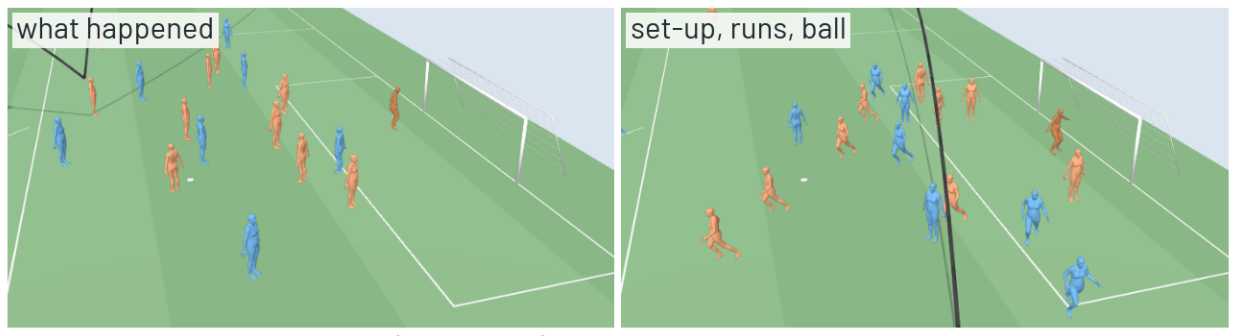

Juventus v Como, 5', corner (Cambiaso): set-up, runs and ball +0.038 xG

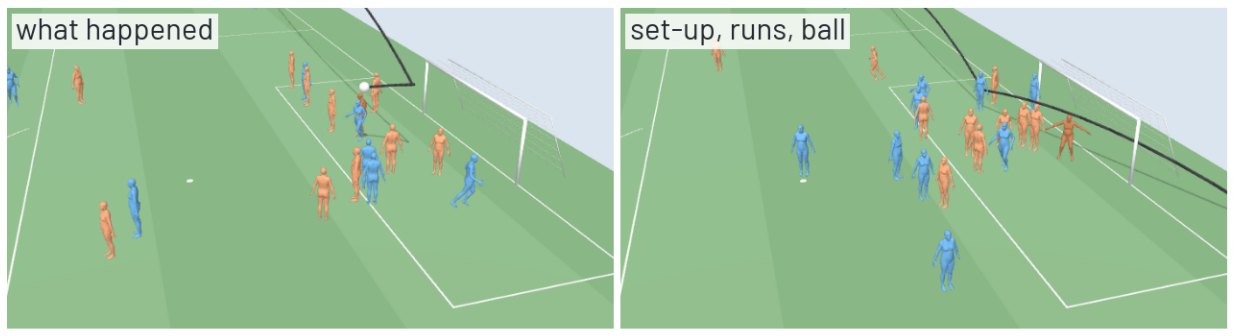

Milan v Roma, 28', corner (Modrić): set-up, runs and ball +0.036 xG

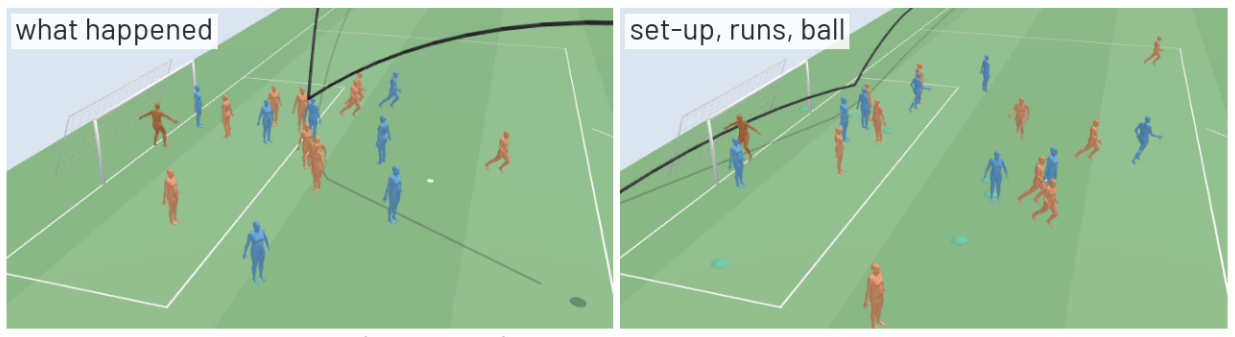

Roma v Como, 42', corner (Pellegrini): set-up, runs and ball +0.036 xG

**Largest examples: gain from the ball, real ball beats the search, what happened far above / below its futures**

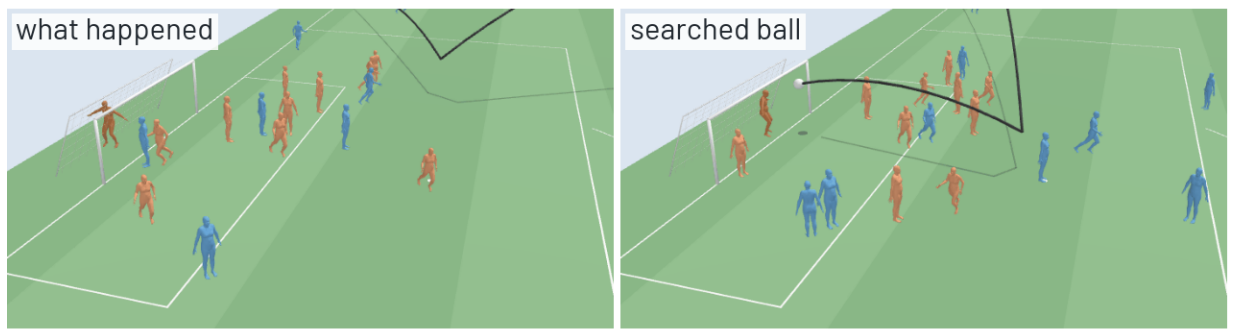

Como v Napoli, 46', corner (Da Cunha): searched ball +0.044 xG (the largest gain)

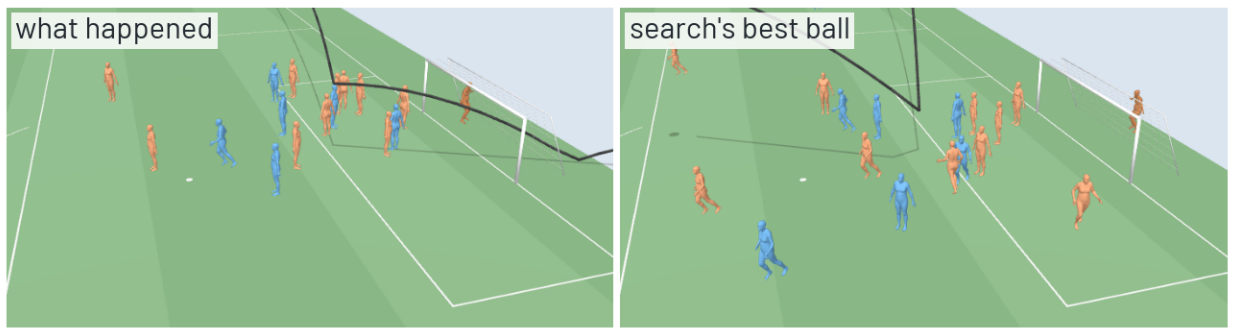

Inter v Juventus, 76', corner (Dimarco): real ball's surplus over the search's best +0.011 xG

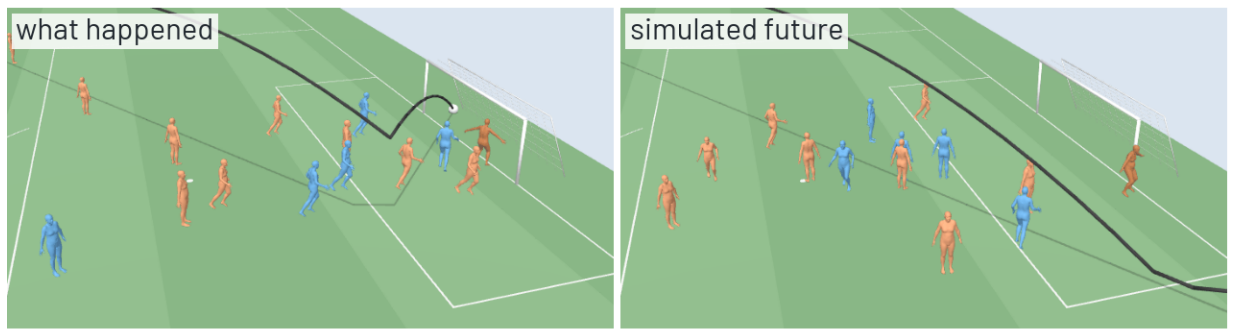

Inter v Como, 58', free kick (Çalhanoğlu): what happened, mean surplus +0.119 xG

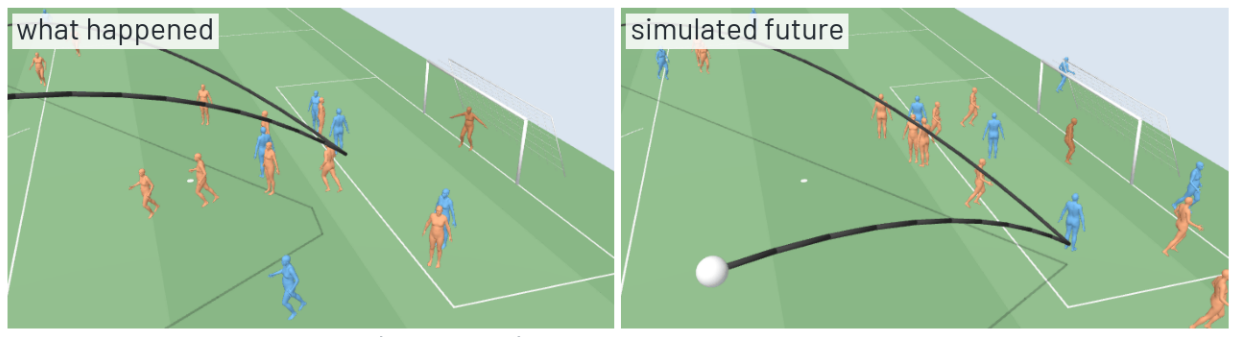

Roma v Inter, 20', free kick (Pellegrini): what happened, mean surplus −0.036 xG

**Figure 7:** Set-up, runs and ball searched together, and the largest examples: the largest surplus of a searched ball, the real ball with the largest surplus over the search's best, and what happened with the highest and lowest mean surplus over its simulated futures.